%% file: main.tex
\documentclass[11pt,a4paper]{article}
\usepackage[hyperref]{acl2021}
\usepackage{times}
\usepackage{latexsym}

\input{math_commands.tex}

\usepackage{url}
\usepackage{algorithm}
\usepackage{enumitem}
\usepackage{algorithmic}
\usepackage{booktabs}
\usepackage{graphicx}
\usepackage{microtype}
\usepackage{colonequals}
\usepackage{amsmath}
\usepackage{amssymb}
\usepackage{mathtools}
\usepackage{xcolor,soul}
\usepackage{comment}

\aclfinalcopy % Uncomment this line for the final submission
\def\aclpaperid{4041} %  Enter the acl Paper ID here

\DeclarePairedDelimiterX{\infdivx}[2]{(}{)}{%
  #1\;\delimsize\|\;#2%
}
\newcommand{\infdiv}{D_{KL}\infdivx}

\title{Learning to Perturb Word Embeddings for Out-of-distribution QA}

\author{Seanie Lee$^1$\thanks{* Equal contribution} \: Minki Kang$^{1}$$^*$ \: Juho Lee$^1$, \: Sung Ju Hwang$^1$$^,$$^2$ \\  
	KAIST$^1$,  AITRICS$^2$,  South Korea\\
	\texttt{\{lsnfamily02, zzx1133, juholee, sjhwang82\}@kaist.ac.kr}}
	
\date{}

\begin{document}
\maketitle
\input{0abstract}
\input{1introduction}
\input{2relatedwork}
\input{3method}
\input{4experiment}
\input{5conclusion}
\input{6broader-impact}

\bibliographystyle{acl_natbib}
\bibliography{main}

\clearpage
\input{7appendix}

\end{document}

%% file: math_commands.tex
\usepackage{amsmath,amsfonts,bm}

\def\eqref#1{equation~\ref{#1}}
\def\1{\bm{1}}

\def\rvc{{\mathbf{c}}}

\def\rve{{\mathbf{e}}}

\def\rvh{{\mathbf{h}}}

\def\rvw{{\mathbf{w}}}
\def\rvx{{\mathbf{x}}}
\def\rvy{{\mathbf{y}}}
\def\rvz{{\mathbf{z}}}

\DeclareMathAlphabet{\mathsfit}{\encodingdefault}{\sfdefault}{m}{sl}
\SetMathAlphabet{\mathsfit}{bold}{\encodingdefault}{\sfdefault}{bx}{n}

\DeclareMathOperator*{\argmax}{arg\,max}

%% file: 0abstract.tex
\begin{abstract}
QA models based on pretrained language models have achieved remarkable performance on various benchmark datasets. However, QA models do not generalize well to unseen data that falls outside the training distribution, due to distributional shifts. Data augmentation (DA) techniques which drop/replace words have shown to be effective in regularizing the model from overfitting to the training data. Yet, they may adversely affect the QA tasks since they incur semantic changes that may lead to wrong answers for the QA task. To tackle this problem, we propose a simple yet effective DA method based on a stochastic noise generator, which learns to perturb the word embedding of the input questions and context without changing their semantics. We validate the performance of the QA models trained with our word embedding perturbation on a single source dataset, on five different target domains. The results show that our method significantly outperforms the baseline DA methods. Notably, the model trained with ours outperforms the model trained with more than 240K artificially generated QA pairs.
\end{abstract}

%% file: 1introduction.tex
\section{Introduction}
% data augmentation is critical for machine learning models - prone to overfitting, requires large amount of data
Deep learning models have achieved impressive performances on a variety of real-world natural language understanding tasks such as text classification, machine translation, question answering, and text generation to name a few~\cite{transformer, bidaf}. Recently, language models that are pretrained with a large amount of unlabeled data have achieved breakthrough in the performance on these downstream tasks~\cite{BERT}, even surpassing human performance on some of them.% and even they have achieved human-level performance in some tasks. 

The success of such data-driven language model pretraining heavily depends on the amount and diversity of training data available, since when trained with a small amount of highly-biased data, the pretrained models can overfit and may not generalize well to out-of-distribution data. Data augmentation (DA) techniques~ \cite{ImageNet, mixup, cutmix, word-dropout} can prevent this to a certain extent, but most of them are developed for image domains and are not directly applicable to augmenting words and texts. Perhaps the most important desiderata for an augmentation method in supervised learning, is that it should not change the label of an example. For image domains, there exist several well-defined data augmentation techniques that can produce diverse augmented images without changing the semantics. In contrast, for Natural Language Processing (NLP), it is not straightforward to augment the input texts without changing their semantics. A simple augmentation technique that preserves semantics is replacing words with synonyms or using back translation~\citep{word-dropout}. However, they do not effectively improve the generalization performance because the diversity of viable transformations with such techniques is highly limited~\citep{meta-backtranslation}.

\begin{comment}
However, what is also crucial to the success of the recent data-driven models is the availability of a large scale labeled dataset. Without a sufficient amount of data, even the pretrained deep learning models are prone to overfitting and fragile to out-of-domain data which falls out of the training distribution. Several data augmentation methods \cite{ImageNet, mixup, cutmix, word-dropout} are known to be effective regularizer so that the model becomes robust to distributional shift. % to prevent the overfitting

For images, there are several well-defined data augmentation methods that preserve the semantics of the original inputs such as rotation, translation, or random cropping~\citep{ImageDataAug}. In contrast, it is not trivial to transform the input without changing the label of the input. In natural language processing (NLP), we increase the number of training data by replacing a word with a synonym or using back translation \cite{word-dropout}. Even though those operations may preserve the semantics of the original input, they may not improve the generalization of the model because the variation of viable transformation is limited \cite{meta-backtranslation}.
\end{comment}

\begin{figure*}[ht!]
    \centering
    \includegraphics[width=1.0\linewidth]{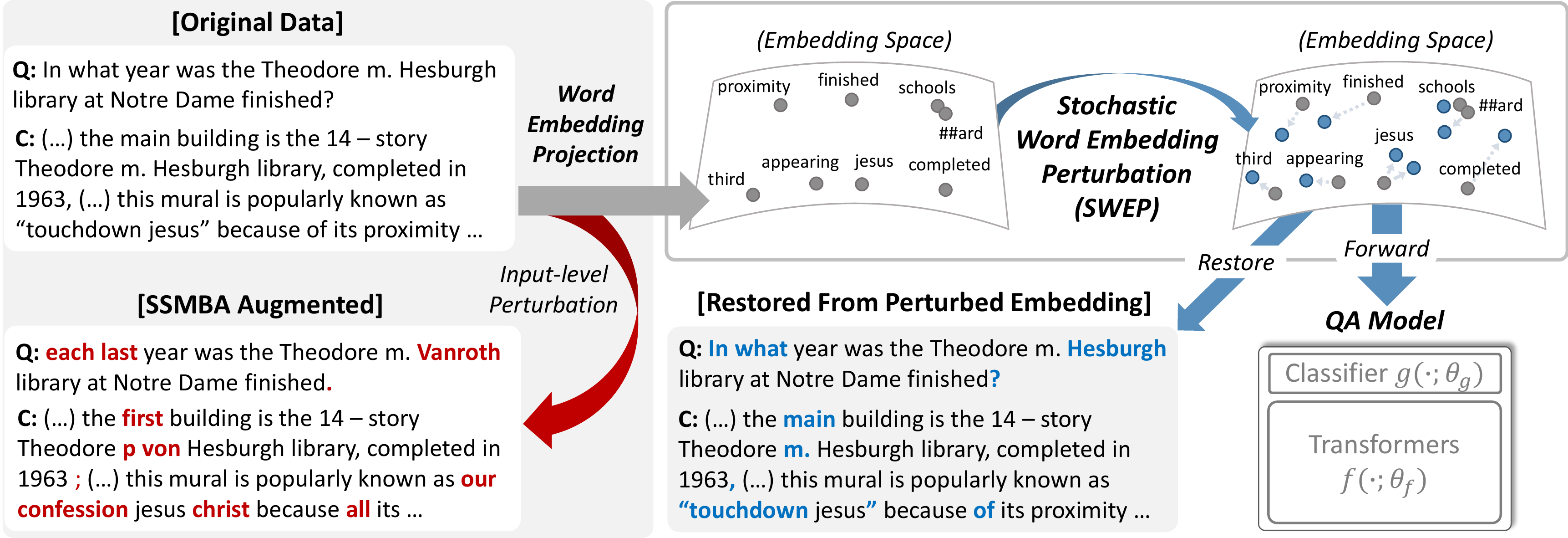}
    \vspace{-0.3in}
    \caption{\small\textbf{Concept.} Our model SWEP perturbs word embedding and feeds the perturbed embedding to the QA model. While the input-level perturbation method (SSMBA) changes the words a lot, our method preserves the original words if we project perturbed embedding back to the words.}
    \label{fig:concept}
    \vspace{-0.2in}
\end{figure*}

Some recent works~\citep{EDA, SSMBA} propose data augmentation methods tailored for NLP tasks based on dropping or replacing words and show that such augmentation techniques improve the performance on the out-of-domain as well as the in-domain tasks. As shown in Fig. \ref{fig:concept}, however, we have observed that most existing data augmentation methods for NLP change the semantics of original inputs. While such change in the semantics may not be a serious problem for certain tasks, it could be critical for \emph{Question Answering} (QA) task since its sensitivity to the semantic of inputs. For instance, replacing a single word with a synonym (\textcolor{blue}{Hesburgh} $\rightarrow$ \textcolor{red}{Vanroth} in Fig. \ref{fig:concept}) might cause the drastic semantic drift of the answer~\citep{adversarial-squad}. Thus, word-based augmentations are ineffective for QA tasks, and most existing works on data augmentation for QA tasks resort to question or QA-pair generation. Yet, this approach requires a large amount of training time, since we have to train a separate generator, generate QA pairs from them, and then use the generated pairs to train the QA model. Also, QA-pair generation methods are not sample-efficient since they usually require a large amount of generated pairs to achieve meaningful performance gains.

% Introduce our method.
To address such limitations of the existing data augmentation techniques for QA, we propose a novel DA method based on learnable word-level perturbation, which effectively regularizes the model to improve its generalization to unseen questions and contexts with distributional shifts. Specifically, we train a stochastic perturbation function to learn how to perturb each word embedding of the input without changing its semantic, and augment the training data with the perturbed samples. We refer to this data augmentation method as Stochastic Word Embedding Perturbation (SWEP).

%\textbf{P}erturbation \textbf{o}f \textbf{W}ord embedding for \textbf{D}ata \textbf{A}ugmentation (POWDA). 

The objective of the noise generator is to maximize the log-likelihood of the answer of the input with perturbation, while minimizing the Kullback-Leibler (KL) divergence between prior noise distribution and conditional noise distribution of given input. Since the perturbation function maximizes the likelihood of the answer of the perturbed input, it learns how to add noise without changing the semantics of the original input. Furthermore, minimizing the KL divergence prevents generating identical noise as the variance of the prior distribution is non-zero, i.e. we can sample diverse noise for the same input. 

% More details on the method needed
We empirically validate our data augmentation method on both extractive and generative QA tasks. We train the QA model on the SQuAD dataset \citep{SQuAD} with our learned perturbations, and evaluate the trained model on the five different domains --- BioASQ \cite{bioasq}, New York Times, Reddit post, Amazon review, and Wikipedia \cite{natural-shift} as well as SQuAD to measure the generalization performance on out-of-domain and in-domain data. The experimental results show that our method improves the in-domain performance as well as out-of-domain robustness of the model with this simple yet effective approach, while existing baseline methods often degrade the performance of the QA model, due to semantics changes in the words. Notably, our model trained only with the SQuAD dataset shows even better performance than the model trained with 240,422 synthetic QA pairs generated from a question generation model. Our contribution in this work is threefold.
% [itemsep=0.25mm, parsep=0pt, leftmargin=*]
\begin{itemize}
\item We propose a simple yet effective data augmentation method to improve the generalization performance of pretrained language models for QA tasks. 

\item We show that our learned input-dependent perturbation function transforms the original input without changing its semantics, which is crucial to the success of DA for question answering.

\item  We extensively validate our method for domain generalization tasks on diverse datasets, on which it largely outperforms strong baselines, including a QA-pair generation method.
\end{itemize}

%% file: 2relatedwork.tex
\section{Related Work}
\paragraph{Data Augmentation}
As in image domains~\citep{ImageNet, AdvAug, cutmix}, data augmentation methods are known to be an effective regularizer in text domain \cite{word-dropout}. However, unlike the image transformations that do not change their semantics, transforming raw texts without changing their semantics is difficult since they are composed of discrete tokens. The most common approach for data augmentation in NLP is applying simple perturbations to raw words, by either deleting a word or replacing it with synonyms~\citep{EDA}. In addition, back-translation with neural machine translation has also been shown to be effective, as it paraphrases the original sentence with a different set and ordering of words while preserving the semantics to some extent~\citep{UDA}. Beyond such simple heuristics, \citet{SSMBA} propose to mask the tokens and reconstruct them with pretrained language model to augment training data for text classification and machine translation. For QA tasks, question or QA-pair generation~\citep{semantic-drift, hcvae} are also popular augmentation techniques, which generate questions or question-answer pairs from an unlabeled paragraph, thus they can be utilized as additional data to train the model.

\paragraph{Domain Generalization}
Unlike domain adaptation in which the target domains are fixed and we can access unlabeled data from them, domain generalization aims to generalize to unseen target domains without access to data from the target distribution. Several prior works \citep{l2g, meta-reg, cross-domain} propose meta-learning frameworks to tackle domain generalization, focusing on image domains. For extractive QA, \citet{robustqa} leverage adversarial training to learn a domain-invariant representation of question and context. However, they require multiple heterogeneous source datasets to train the model to be robust to Out-of-Domain data. In contrast, \citet{AdvAug} leverage adversarial perturbation to generate fictitious examples from a single source dataset, that can generalize to unseen domains. 

%% file: 3method.tex
\section{Method}
\subsection{Brief Summary of Backgrounds}
The goal of extractive Question Answering (QA) is to point out the start and end position of the answer span $\rvy = (y_{\text{start}}, y_{\text{end}})$ from a paragraph (context) $\rvc = (c_1, \ldots, c_L)$ with length $L$ for a question $\rvx =(x_1, \ldots, x_M)$.  For generative QA, it aims to generate answer $\rvy = (y_1, \ldots, y_K)$ instead of predicting the position of answer spans from the context. A typical approach to the QA is to train a neural networks to model the conditional distribution $p_\theta(\rvy| \rvx, \rvc)$, where $\theta$ are composed of $\theta_f \text{ and } \theta_g$ denoted for the parameters of the encoder $f(\cdot;\theta_f)$ and classifier or decoder $g(\cdot;\theta_g)$  on top of the encoder. We estimate the parameter $\theta$ to maximize the log likelihood with $N$ observations $\{\rvx^{(i)}, \rvy^{(i)}, \rvc^{(i)}\}_{i=1}^N$, which are drawn from some unknown distribution $p_\text{train}$, as follows:
\begin{equation}
    \mathcal{L}_{MLE}(\theta) \coloneqq \sum_{i=1}^N \log p_\theta(\rvy^{(i)}| \rvx^{(i)}, \rvc^{(i)})
\label{erm}
\end{equation}
For convenience, we set the length $T \coloneqq L+M+3$ and abuse notations to define the concatenated sequence of the question $\rvx$ and context $\rvc$ as $\rvx \coloneqq (x_0,  \ldots, x_L, c_0, \ldots, c_{M+1})$ where $x_0, c_0, c_{M+1}$ denote start, separation, and end symbol, respectively.

However, the model trained to maximize the likelihood in Eq. (\ref{erm}) is prone to overfitting and brittle to distributional shifts where target distribution $p_\text{test}$ is different from $p_\text{train}$. In order to tackle this problem, we train the model with additional data drawn from different generative process to increase the support of training distribution, to achieve better generalization on novel data with distributional shifts. We will describe it in the next section.

\subsection{Learning to Perturb Word Embeddings}
\label{sec-3.2}
Several methods for data augmentation have been proposed in text domain, however, unlike in image domains~\cite{mixup, manifold-mixup, cutmix}, there does not exist a set of well-defined data augmentation methods which transform the input without changing its semantics. We propose a new data augmentation scheme where we sample a noise $\rvz = (\rvz_1, \ldots, \rvz_T)$ from a distribution $q_\phi (\rvz| \rvx)$ and perturb the input $\rvx$ with the sampled noise without altering its semantics. To this end, the likelihood $p_\theta (\rvy|\rvx,\rvz)$ should be kept high even after the perturbation, while the perturbed instance should not collapse to the original input. We estimate such parameters $\phi$ and $\theta$ by maximizing the following objective:
\begin{align}
\begin{split}
    \mathcal{L}_{noise}(\phi, \theta) &\coloneqq \sum_{i=1}^N \mathbb{E}_{q_\phi(\rvz|\rvx^{(i)})}[\log p_\theta (\rvy^{(i)}|\rvx^{(i)},\rvz)] \\
    &- \beta\sum_{t=1}^T \infdiv{q_\phi (\rvz_t|\rvx^{(i)})}{p_\psi (\rvz_t)}
\label{noise:eq}
\end{split}
\end{align}
where $\beta \geq 0$ is a hyper-parameter which controls the effect of KL-term. We assume that $\rvz_{t}$ and $\rvz_{t'}$ are conditionally independent given $\rvx$ if $t\neq t'$, i.e., $q_\phi (\rvz|\rvx) = \prod_{t=1}^T q_\phi (\rvz_t|\rvx)$. The parameter of prior $\psi$ is a hyper-parameter to be specified. When $\beta = 1$, the objective corresponds to the Evidence Lower BOund (ELBO) of the marginal likelihood.

Maximizing the expected log-likelihood term in Eq. (\ref{noise:eq}) increases the likelihoods evaluated with the perturbed embeddings, and therefore the semantics of the inputs after perturbations are likely to be preserved. 
The KL divergence term in Eq. (\ref{noise:eq}) penalizes the perturbation distribution  $q_\phi(\rvz|\rvx)$ deviating too much from the prior distribution $p_\psi(\rvz)$. We assume that the prior distribution is fully factorized, i.e. $p_\psi(\rvz_1, \ldots, \rvz_T)=\prod_{t=1}^T p_\psi (\rvz_t)$. Furthermore, we set each distribution $p_\psi (\rvz_t)$ as a multivariate Gaussian distribution $\mathcal{N}(\mathbf{1}, \alpha \textbf{I}_d)$, where $\mathbf{1} = (1,\ldots, 1) \in \mathbb{R}^d$, $\textbf{I}_d, \alpha$ denotes a vector with ones, identity matrix, and positive real number, respectively. Hence, we expect the inputs perturbed with the multiplicative noises remain close to the original inputs on average. Note that the choice of the prior is closely related to Gaussian dropout~\citep{dropout}; we will elaborate on this connection later.

The parameterization of the perturbation function $q_\phi$ heavily affects the success of the learning with the objective (\ref{noise:eq}). The function needs to control the intensity of perturbation for each token of $\rvx$ without changing the semantics. Since the meaning of each word varies across linguistic contexts, the function should be expressive enough to encode the sentence $\rvx$ into a meaningful latent space embedding to contextualize the subtle meaning of each word in the sentence.

\begin{figure}
    \centering
    \includegraphics[width=1.0\linewidth]{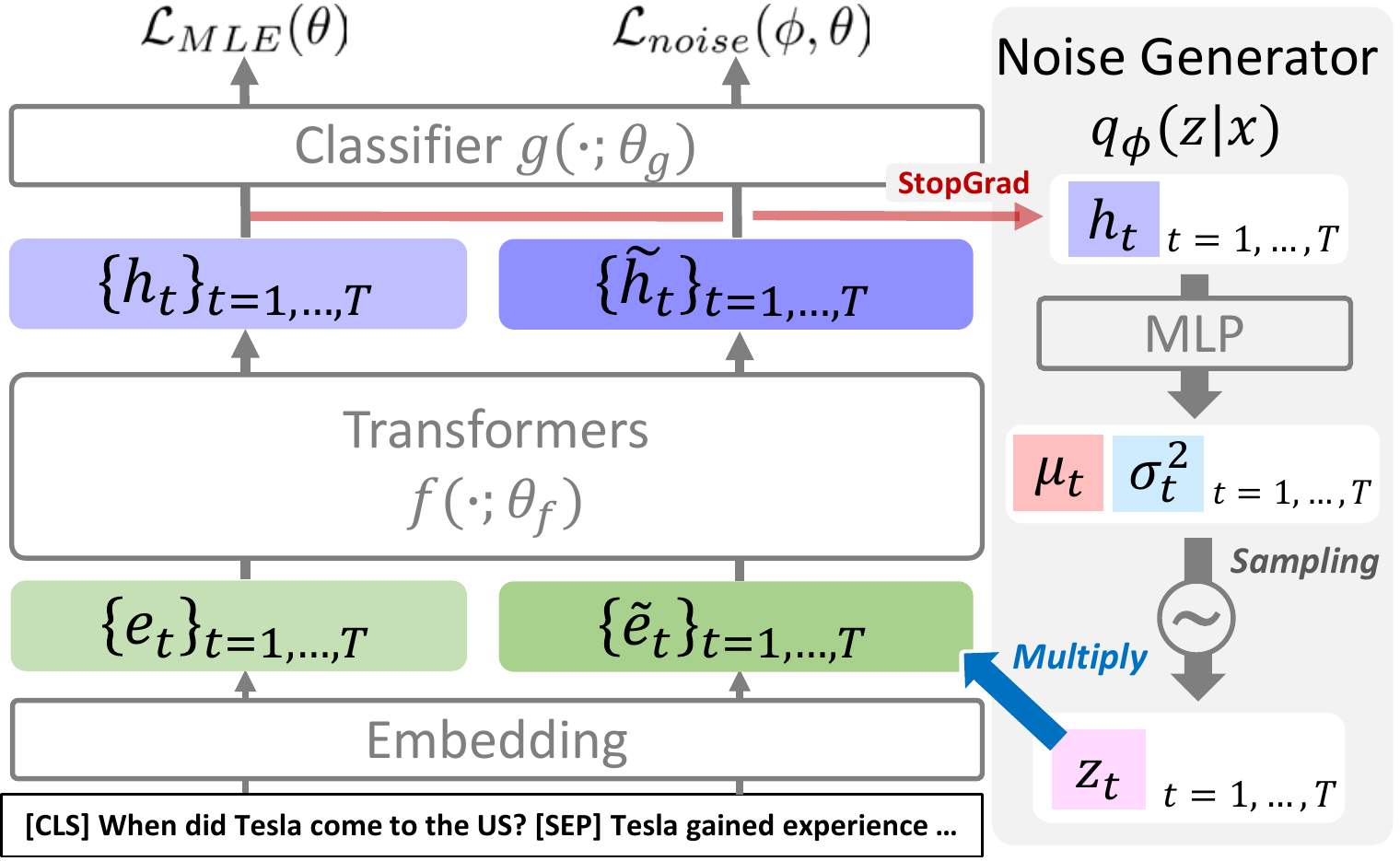}
    \vspace{-0.2in}
    \caption{\textbf{Architecture.} Overview of how the input is perturbed with SWEP. It encodes the input to hidden representation with transformers and outputs a desirable noise for each word embedding. The noise is multiplied with the word embedding.}
    \vspace{-0.2in}
    \label{fig:arch}
\end{figure}

To this end, we share the encoder function $f(\cdot; \theta_f)$ to contextualize the input $\rvx$ into hidden representation $(\rvh_1, \ldots, \rvh_T)$ and feed it into the perturbation function as input as shown in the left side of Fig. \ref{fig:arch}. However, we stop the gradient of $\phi$ with respect to $\mathcal{L}(\phi, \theta)$ propagating to the encoder $f(\cdot; \theta_f).$ Intuitively, it prevents noisy gradient from flowing to $p_\theta$ for early stage of training. On top of the encoder, we stack two layer feed forward neural network with ReLU activation, which outputs mean $\boldsymbol{\mu}_t \in \mathbb{R}^d$ and variance $\boldsymbol{\sigma}^2_t \in \mathbb{R}^d$ for each token, following \citet{vae}. We leverage the reparameterization trick \cite{vae} to sample $\rvz_t \in \mathbb{R}^d$.  Since $\rvx$ is a sequence of discrete tokens, we map each token $x_t$ to corresponding word embedding $\rve_t$ and multiply it with the noise $\rvz_t$ in element-wise manner as follows:
\begin{equation}
\begin{gathered}
        \rve_t = \text{WordEmbedding}(x_t) \\
        (\rvh_1, \ldots, \rvh_T) = f(\rve_1, \ldots, \rve_T;\theta_f) \\
        \boldsymbol{\mu}_t, \boldsymbol{\sigma}^2_t = \text{MLP}(\rvh_t) \\
        \rvz_t = \boldsymbol{\mu}_t + \boldsymbol{\sigma}_t \odot \epsilon \text{, where } \epsilon \sim \mathcal{N}(\mathbf{0}, \mathbf{I}_d) \\
        \Tilde{\rve}_t = \rve_t \odot \rvz_t
\label{powda:eq}
\end{gathered}
\end{equation}
where $\odot$ denotes element-wise multiplication.
We  feed $(\Tilde{\rve}_1, \ldots, \Tilde{\rve}_T)$ to the $g \circ f$ to compute the likelihood $p_\theta (\rvy|\rvx, \rvz)$ as shown in Fig. \ref{fig:arch}. 

\subsection{Learning Objective}
As described in the section \ref{sec-3.2}, we can jointly optimize the parameters $\theta, \phi$ with gradient ascent. However, we want to train the QA model with additional data drawn from the different generative process as well as the given training data to increase the support of training distribution, which leads to better regularization and robustness to the distributional shift. Therefore, our final learning objective function is a convex combination of $\mathcal{L}_{MLE}(\theta)$ and $\mathcal{L}_{noise}(\phi, \theta)$ as follows:
\begin{equation}
    \mathcal{L}(\phi, \theta) = \lambda \mathcal{L}_{MLE}(\theta) +  (1-\lambda)\mathcal{L}_{noise}(\phi, \theta)
\label{obj}    
\end{equation}
where $0 < \lambda <1$ is a hyper-parameter which controls the importance of each objective. For all the experiments, we set $\lambda$ as 0.5. In other words, we train the QA model to maximize the conditional log-likelihood of the original input and perturbed one with stochastic gradient ascent. 

\subsection{Connection to Dropout}
Since each random variable of the perturbation vector $\rvz_t=(z_{t,1}, \ldots, z_{t,d})$ is independent, we only consider the $i-$th coordinate. With the reparameterization trick, we can write $z_{t,i} = \mu_{t,i} + \sigma_{t,i} \odot \epsilon$, where each $\epsilon \stackrel{iid}{\sim} \mathcal{N}(0,1)$ and $\mu_{t,i}, \sigma_{t,i}$ are $i-$th component of $\boldsymbol{\mu}_t, \boldsymbol{\sigma}_t$ which are outputs of neural network as described in Eq. (\ref{powda:eq}). Simply, each noise element $z_{t,i}$ is sampled from $\mathcal{N}(\mu_{t,i}, \sigma_{t,i}^2)$. Assume that $\tilde{z}$ is the noise sampled from the  prior distribution $\mathcal{N}(1, \alpha)$, i.e.  $\tilde{z} = 1 + \sqrt{\alpha} \cdot \epsilon$. Then, $z_{t,i}$ can be expressed in terms of $\tilde{z}$ as follows:
\begin{equation}
    z_{t,i} = \frac{\sigma_{t,i}}{\sqrt{\alpha}}\tilde{z} + (\mu_{t,i} - \frac{\sigma_{t,i}}{\sqrt{\alpha}})
\end{equation}
If we set $\alpha = (1-p)/p$ where $p$ is the retention probability, we can consider $\Tilde{z}$ as a Gaussian dropout mask sampled from $\mathcal{N}(1, \frac{1-p}{p})$, which shows comparable performance to dropout mask sampled from Bernoulli distribution with probability $p$ \cite{dropout}. Then, we can interpret our perturbation function as the input dependent dropout which scales and translates the Gaussian dropout mask, and thus it flexibly controls the intensity of perturbation adaptively to each word embedding of the input $\rvx$.

\begin{table*}
	\small
	\centering
	\begin{tabular}{lcccccc}
		\toprule
		{\textbf{Method}} & \textbf{SQuAD} & {\textbf{Wiki}} &  {\textbf{NYT}} &  {\textbf{BioASQ}} & {\textbf{Reddit}} & {\textbf{Amazon}}   \\
		\midrule[0.8pt]
		\multicolumn{7}{c}{\textbf{BERT-base-uncased} (EM / F1) } \\ 
		\midrule[0.8pt]
		\textbf{MLE} & {81.32 / 88.62} & {76.42 / 87.02 } & {77.54 / 86.54 } & {45.34 / 59.77} & {63.94 / 76.97} & {60.74 / 75.38} \\
		\textbf{Adv-Aug} & {81.39 / 88.71} &  {77.29 / 88.38} & {77.67 / 86.53} & {45.47 / 60.30} & {64.55 / 77.61} & {61.38 / 75.83} \\
		\textbf{Word-Dropout} & {81.03 / 88.21} & {76.94 / 87.30} & {76.67 / 85.99} & {44.34 / 58.93} & {65.05 / 77.96} & {60.87 / 75.71} \\
		\textbf{Gaussian-Dropout} & {81.47 / 88.78} & {77.28 / 87.23} & {77.25 / 86.35} & {45.27 / 61.37} & {65.19 / 77.73} & {61.67 / 75.98}\\
		\textbf{Bernoulli-Dropout} & {81.46 / 88.76} & {77.34 / 87.40}& {77.16 / 86.35} & {44.21 / 59.33} & {64.53 / 77.25} & {61.27 / 75.85} \\
		\textbf{SSMBA} & {78.17 / 86.53} & {74.33 / 85.26} & {74.31 / 83.98} & {39.96 / 54.49} & {59.29 / 73.50} & {56.57 / 71.81}  \\
		\midrule[0.5pt]
		\textbf{Prior-Aug} &{81.77 / 89.04} & {77.95 / 87.83} & {77.92 / 86.81} & {46.40 / 60.80} & {65.50 / 78.16} & {61.57 / 76.22} \\
		\textbf{SWEP} & \textbf{82.24 / 89.43} & \textbf{78.60 / 88.28} & \textbf{78.11 / 86.92} & \textbf{47.27 / 61.72} & \textbf{65.93 / 78.45} & \textbf{62.42 / 76.84}\\
		\midrule[0.8pt]
		\multicolumn{7}{c}{\textbf{ELECTRA-small-uncased} (EM / F1) } \\
		\midrule[0.8pt]
		\textbf{MLE} & {76.95 / 84.92} & {73.57 / 84.30} & {73.68 / 82.93} & {38.63 / 54.32} & {59.59 / 72.33} & {57.93 / 72.06} \\
		\textbf{Adv-Aug} & {75.81 / 84.40} & {73.69 / 84.23} & {73.37 / 82.89} & {38.23 / 53.4} & {\textbf{59.97} / 73.33} & {59.44 / 73.36} \\
		\textbf{Word-Dropout} & {75.81 / 84.19} & {72.94 / 83.90} & {72.96 / 82.24} & {39.29 / 54.02} & {59.04 / 72.12} & {58.49 / 72.41} \\
		\textbf{Gaussian-Dropout} & {76.42 / 84.53}  & {73.31 / 84.11} & {73.27 / 82.51} & {37.30 / 52.46} & {59.29 / 72.31} & {57.50 / 71.65}\\
		\textbf{Bernoulli-Dropout} & {76.31 / 84.50} & {73.50 / 84.08} & {73.35  / 82.75} & {37.10 / 52.37} & {59.33 / 72.56} & {57.71 / 71.99} \\
		\textbf{SSMBA} & {77.75 / 85.81} & {\textbf{74.90} / \textbf{85.21}} & {73.25 / 82.62} &{39.02 / 53.32} & {58.97 / 72.83} & {56.66 / 71.89}  \\
		\midrule[0.5pt]
		\textbf{Prior-Aug} & {77.70 / 85.60} & {74.65 / 85.02} & {74.38 / 83.47} & {38.96 / 54.19} & {59.92 / 73.10} & {59.01 / 73.11} \\
		\textbf{SWEP} & \textbf{77.78 / 85.86} & 74.25 / 85.20 & \textbf{75.18 / 84.18} & \textbf{40.35 / 55.72} & {59.68 /\textbf{ 73.97}} & \textbf{60.89 / 74.06} \\
		\bottomrule
	\end{tabular}
	\vspace{-0.1in}
	\caption{\small Experimental results of extractive QA with BERT and ELECTRA model on six different test dataset.}
	\label{qa-exp}
	\vspace{-0.15in}
\end{table*}

%% file: 4experiment.tex
\section{Experiment}
\subsection{Task}
Our goal is to regularize the QA model to generalize to unseen domains, such that it is able to answer the questions from the new domain. We consider a more challenging setting where the model is trained with a single source dataset and evaluate it on the datasets from the unseen domains as well as on unseen examples from the source domain. Specifically, we train the QA model with SQuAD dataset \cite{SQuAD} as source domain, test the model with several different target domain QA datasets --- BioASQ \cite{bioasq}, New Wikipedia (Wiki), New York Times (NYT), Reddit posts, and Amazon Reviews \cite{natural-shift}. We evaluate the QA model with F1 and Exact Match (EM) score, following the convention for extractive QA tasks. For the BioASQ dataset, we use the dataset provided in the MRQA shared task \cite{mrqa}. We downloaded the other datasets from the official website of \citet{natural-shift}.
\label{task}

\subsection{Experimental Setup}
\paragraph{Implementation Detail} As for the encoder $f$, we use the pretrained language model --- BERT-base \cite{BERT}, ELECTRA-small \cite{Electra} for extractive QA and randomly initialize an affine transformation layer for $g$. For the generative QA task, we use a T5-small \cite{T5} for $f\circ g$ as an encoder-decoder model. For the perturbation function $q_\phi$, we stack two feed-forward layers with ReLU on the encoder as described in section \ref{sec-3.2}. For the extractive QA task, we train the model for 2 epochs with the batch size 8 and use AdamW optimizer \cite{adamw} with the learning rate $3\cdot 10^{-5}$. For the T5 model, we train it for 4 epochs with batch size 64 and use Adafactor optimizer \cite{adafactor} with learning rate $10^{-4}$. We use beam search with width 4 to generate answers for generative question answering.
\label{setup}

\paragraph{Baselines} We experiment with our model SWEP and its variant against several baselines.
\begin{enumerate}[itemsep=0mm, parsep=0pt, leftmargin=*]
    \item \textbf{MLE}: This is the base QA model fine-tuned to maximize $\mathcal{L}_{MLE}(\theta)$.
    
    \item \textbf{Adv-Aug}: Following \citet{AdvAug}, we perturb the word embeddings of the input $\rvx$ with an adversarial objective and use them as additional training data to maximize $\mathcal{L}_{MLE}(\theta).$ We assume that the answer for each question and context remains the same after the adversarial perturbation. 
    
    \item \textbf{Gaussian-Dropout} This is the model whose word embedding is perturbed with dropout mask sampled from a Gaussian distribution $\mathcal{N}(1, \frac{1-p}{p})$, where $p$ is dropout probability and set to be 0.1 \cite{dropout}.
    
    \item \textbf{Bernoulli-Dropout} This is the model of which word embedding is perturbed with dropout mask sampled from Bernoulli distribution $\text{Ber}(1-p)$, where $p$ is dropout probability and set to be 0.1 \cite{dropout}.
    
    \item \textbf{Word-Dropout}: This is the model trained to maximize $\mathcal{L}_{MLE}(\theta)$ with 
    word dropout \cite{word-dropout} where the tokens of $\rvx$ are randomly set to a zero embedding.
    
    \item \textbf{SSMBA}:  This is the QA model trained to maximize $\mathcal{L}_{MLE}(\theta)$, with additional examples generated by the technique proposed in ~\cite{SSMBA}, which are generated by corrupting the target sequences and reconstructing them using a masked language model, BERT.
    
    \item \textbf{Prior-Aug} This is variant of SWEP trained with additional perturbed data, where the noise is drawn from the prior distribution $p_\psi(\rvz)$ rather than $q_\phi(\rvz|\rvx).$
    
    \item \textbf{SWEP}: This is our full model which maximizes the objective function in Eq. (\ref{obj}).
    
\end{enumerate}

\subsection{Experimental Result} We compare SWEP and its variant Prior-Aug with the baselines as described in section \ref{task}. As shown in Table \ref{qa-exp}, our model outperforms all the baselines, whose backbone networks are BERT or ELECTRA, on most of the datasets. The data augmentation with SSMBA improves the performance of ELECTRA on in-domain dataset SQuAD and Wiki. However, it significantly underperforms ours on out-of-domain datasets even if the data augmentation with SSMBA use 4.8 times more data than ours. Similarly, Table \ref{generative-qa-exp} shows that the T5 model trained with our method consistently improves the performance of the model trained with MLE on most of the datasets. 

Contrary to ours, SSMBA significantly degrades the performance of the BERT and T5 model both on in-domain and out-of-domain datasets. Since masking and reconstructing some of the tokens from a sentence with a masked language model may cause a semantic drift, those transformations make some questions unanswerable. As a result, the data augmentation with SSMBA often hurts the performance of the QA model. 
Similarly, Word-Dropout randomly zeros out word embedding of tokens, but some of zeroed out words are critical for answering questions. Adv-aug marginally improves the performance, but it requires an additional backward pass to compute the gradient for adversarial perturbation, which slows down the training procedure.
\begin{table*}
	\small
	\centering
	\begin{tabular}{lcccccc}
		\toprule
		{\textbf{Method}} & \textbf{SQuAD} & {\textbf{Wiki}} &  {\textbf{NYT}} &  {\textbf{BioASQ}} & {\textbf{Reddit}} & {\textbf{Amazon}}   \\
		\midrule[0.8pt]
		\multicolumn{7}{c}{\textbf{T5-small} (EM / F1) } \\
		\midrule[0.8pt] 
		\textbf{MLE} & {\textbf{77.19} / 85.66} & {72.88 / 84.17} & {75.10 / 83.88} & {40.82 / 54.18} & \textbf{61.19 / 74.25} & {57.52 / 72.16} \\
		\textbf{Adv-Aug} & {74.90 / 84.19} & {71.03 / 82.94} & {73.46 / 82.84} & {38.76 / 52.79} & {58.78 / 72.57} & {54.73 / 70.10} \\
		\textbf{Word-Dropout} & {75.20 / 84.33} & {72.19 / 83.46} & {74.27 / 83.24} & {38.96 / 52.84} & {59.32 / 72.40} & {55.58 / 70.49} \\
		\textbf{Gaussian-Dropout} & {76.25 / 84.86} & {72.56 / 83.69} & {74.76 / 83.57} & {41.15 / 54.64} & {60.14 / 73.40} & {57.01 / 71.52}\\
		\textbf{Bernoulli-Dropout} & {75.15 / 84.34}& {71.64 / 83.33} & {73.81 / 83.06} & {39.42 / 53.77} & {59.06 / 72.48} & {55.22 / 70.46}  \\
		\textbf{SSMBA} & {74.94 / 84.19} & {71.97 / 83.85} & {73.29 / 82.79} & {37.96 / 51.57} & {58.54 / 72.51} & {55.05 / 70.62}  \\
		\midrule[0.5pt]
		\textbf{Prior-Aug} & {76.88 / 85.47} & {73.11 / 84.18} & {75.52 / 84.04} & {40.49 / 54.47} & {60.92 / 74.04} & \textbf{57.99 / 72.38}  \\
		\textbf{SWEP} & {77.12 / \textbf{85.67}} & \textbf{73.34 / 84.35} & \textbf{76.42 / 84.81} & \textbf{43.01 / 55.80} & {60.78 / 73.93} & {57.75 / 72.20} \\
		\bottomrule
	\end{tabular}
	\vspace{-0.1in}
	\caption{\small Experimental results of generative QA with T5-small model on six different test dataset.}
	\label{generative-qa-exp}
	\vspace{-0.2in}
\end{table*}

\begin{figure}
    \centering
    \includegraphics[width=1.0\linewidth]{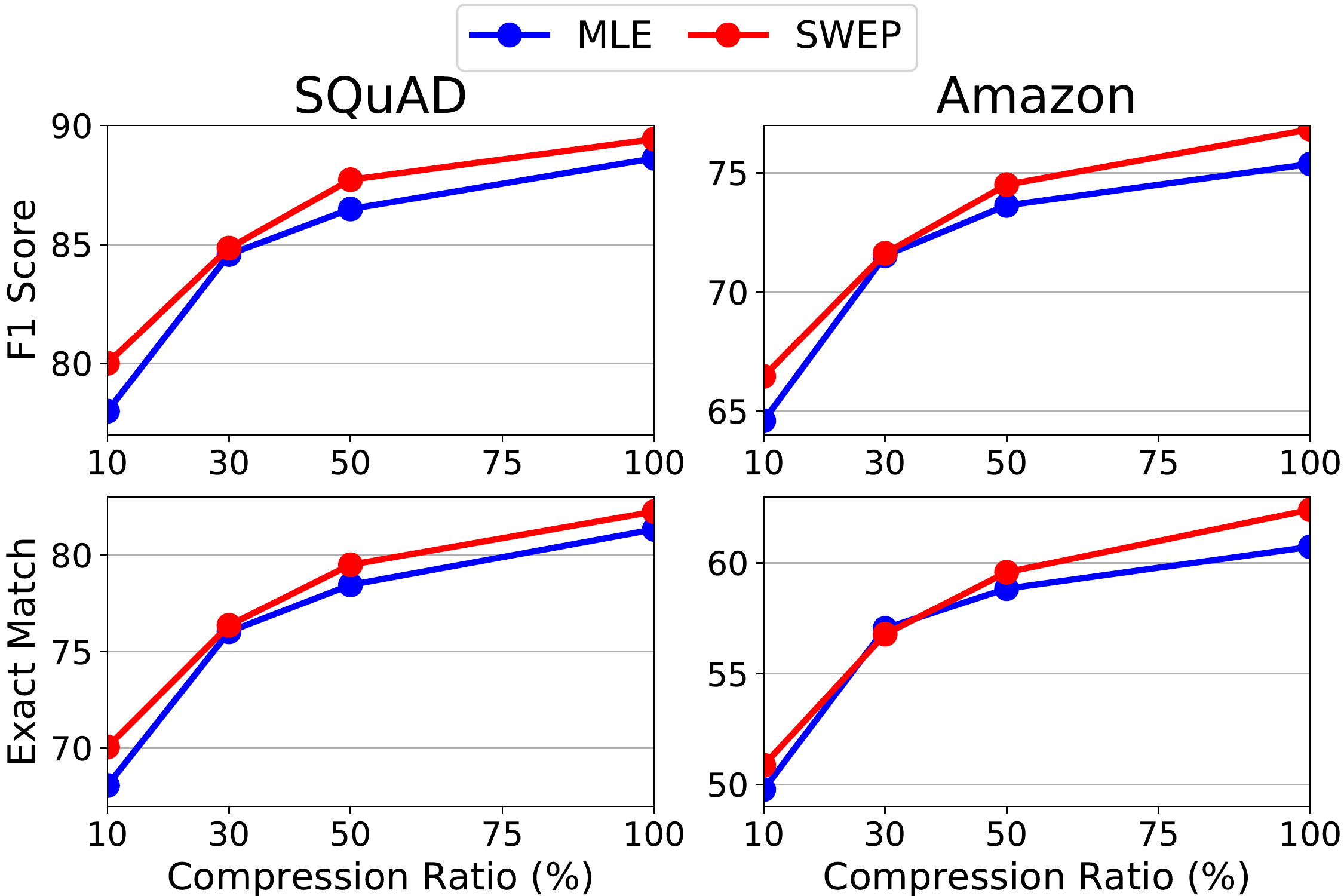}
    \vspace{-0.15in}
    \caption{\small EM or F1 score on SQuAD and Amazon vs. percentage of QA pairs from SQuAD.}
    \label{fig:low-resource}
    \vspace{-0.2in}
\end{figure}

\subsection{Low Resource QA} We empirically show that our data augmentation SWEP is an effective regularizer in the setting where there are only a few annotated training examples. To simulate such a scenario, we reduce the number of labeled SQuAD data to $80\%, 50\%, 30\%, \text{ and } 10\%$ and train the model with the same experimental setup as described in section \ref{setup}. Fig. \ref{fig:low-resource} shows the accuracy as a function of the percentage of QA pairs. Ours consistently improves the performance of the QA model at any ratios of labeled data. Even with $10\%$ of labeled data, it increases EM and F1 score by 1\%.  

\begin{figure}[t!]
    \centering
    \includegraphics[width=1.0\linewidth]{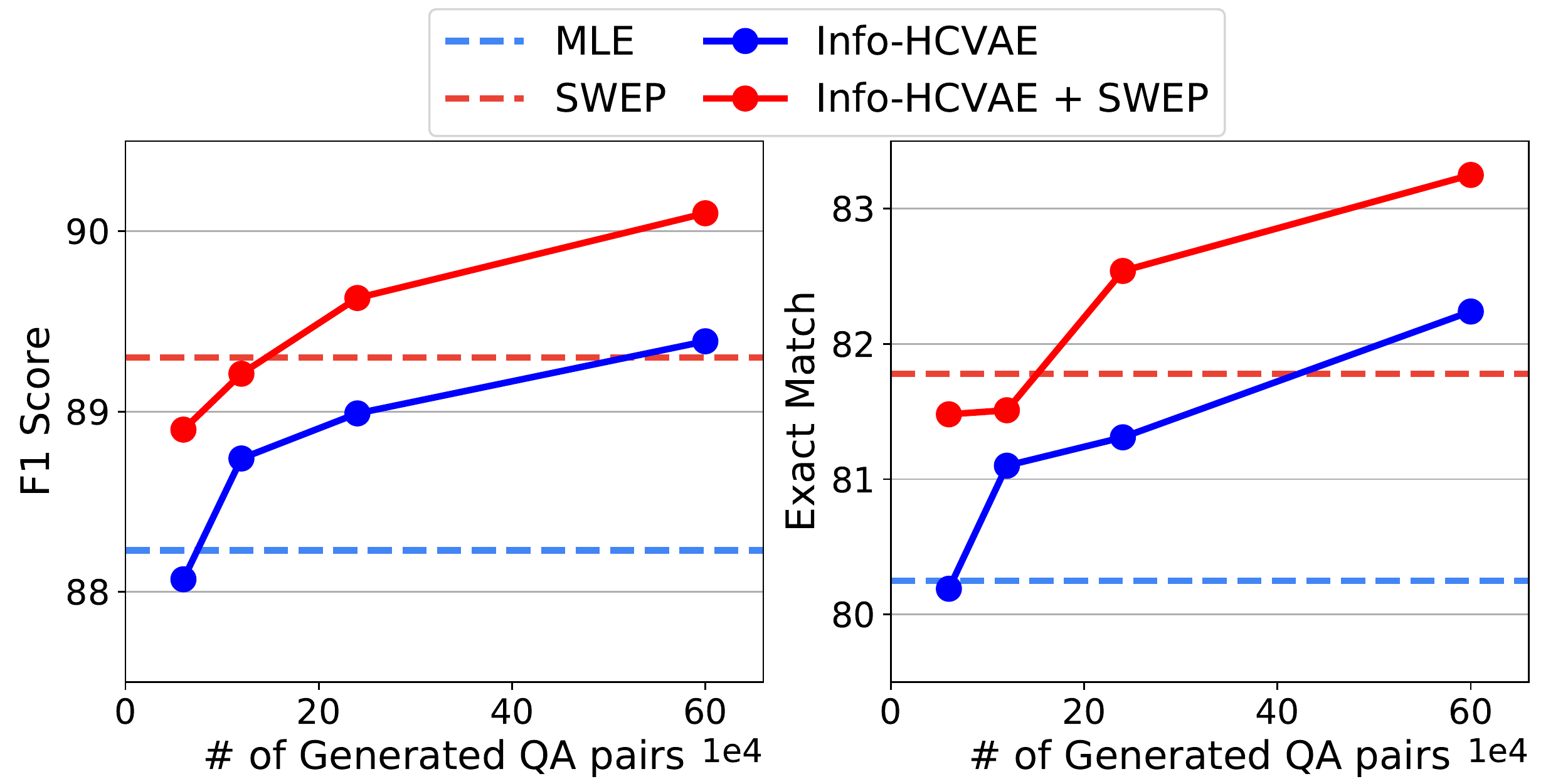}
    \caption{\small EM or F1 score on SQuAD vs. the number of generated QA pairs. Dashed lines indicate results without any synthetic QA pairs.}
    \label{fig:qg}
    \vspace{-0.1in}
\end{figure}

\subsection{Data augmentation with QG} We show that our data augmentation is sample-efficient and further improves the performance of the QA model trained with additional synthetic data generated from the question-answer generation model (QG). We use Info-HCVAE~\citep{hcvae} to generate QA pairs from unlabeled paragraphs and train the BERT model with human-annotated and synthetic QA pairs, while varying the number of the generated pairs. As shown in Fig. \ref{fig:qg}, SWEP trained only with SQuAD already outperforms the model trained with 240,422 synthetic QA pairs generated with Info-HCVAE. Moreover, when combining the two methods, we achieve even larger performance gains compared to when using either SWEP or Info-HCVAE alone, as the two approaches are orthogonal.

\section{Analysis and Discussion}
\subsection{Ablation Study}
We further perform an ablation study to verify the effectiveness of each component of SWEP. In Table~\ref{tab:ablation}, we present the experimental results while removing various parts of our model. 
First of all, we replace the elementwise multiplicative noise with elementwise additive noise and set the prior distribution as $\mathcal{N}(\mathbf{0}, \alpha \textbf{I}_d)$. We observe that the noise generator does not learn meaningful perturbation, which leads to performance degradation.
Moreover, instead of learning $\boldsymbol{\mu}_t$ or $\boldsymbol{\sigma}_t$ from the data, we fix either of them and perform experiments, which we denote w/ fixed $\mu$ and w/ fixed $\sigma$. For all the time step $t$, we set $\boldsymbol{\mu}_t$ as $(1, \ldots, 1) \in \mathbb{R}^d$ for w/ fixed $\mu$. For w/ fixed $\sigma$, we set $\boldsymbol{\sigma}_t^2$ as $(1, \ldots, 1) \in \mathbb{R}^d$, i.e. we use the identity matrix $\textbf{I}_d$ as the covariance of $q_\phi(\rvz|\rvx)$. As shown in Table \ref{tab:ablation}, fixing $\boldsymbol{\mu}_t$ or $\boldsymbol{\sigma}^2_t$ with predefined values achieves slightly better performance than the Prior-Aug, but it degrades the performance of the full model. Based on this experimental results, we verify that learning $\boldsymbol{\mu}_t$ or $\boldsymbol{\sigma}_t^2$ for each word embedding $\rve_t$ is crucial to the success of the perturbation function, as it can delicately perturb each words with more flexibility. 

\begin{table}
\centering
\resizebox{0.48\textwidth}{!}{
    \begin{tabular}{lccc}
    \toprule
    \textbf{ELECTRA-small}  & \textbf{BioASQ}         & \textbf{NYT}           & \textbf{Amazon} \\ 
    \midrule
    Prior-Aug               & {38.96 / 54.19}         & {74.38 / 83.47}        & {59.01 / 73.11} \\
    SWEP                   & \textbf{40.35 / 55.72}  & \textbf{75.18 / 84.18} & \textbf{60.89 / 74.97} \\
    \midrule
    additive perturb.       & {39.16 / 55.15}         & {73.87 / 83.1}        & {59.07 / 73.53} \\
    w/ fixed $\mu = \mathbf{1}$          & {38.36 / 54.29}         & {74.51 / 83.68}        & {59.99 / 73.65}\\
    w/ fixed $\sigma = \textbf{I}_d$       & {38.90 / 54.74}         & {73.34 / 82.79}        & {59.09 / 72.80} \\ 
    w/o $\epsilon \sim \mathcal{N}(\mathbf{0}, \mathbf{I}_d)$ & {38.83 / 54.38}         & {74.62 / 83.60 }        & {59.69 / 73.31} \\
    w/o $D_{KL}$            & {38.90 / 54.65}         & {73.32 / 82.66}        & {59.10 / 72.74} \\ 
    w/o $\:\mathcal{L}_{MLE}(\theta)$ & {37.89 / 53.80} & {72.58 / 82.88} & {58.16 / 72.59}\\
    \bottomrule
    \end{tabular}
}
\vspace{-0.1in}
\caption{\small Ablation study on ELECTRA model.}
\vspace{-0.2in}
\label{tab:ablation}
\end{table}

Furthermore, we convert the stochastic perturbation to deterministic one, which we denote as w/o $\epsilon \sim \mathcal{N}(\mathbf{0}, \textbf{I}_d).$ To be specific, the $\text{MLP}(\rvh_t)$ in Eq. (\ref{noise:eq}) only outputs $\boldsymbol{\mu}_t$ alone and we multiply it with $\rve_t$ without any sampling, i.e. $\Tilde{\rve}_t = \rve_t \odot \boldsymbol{\mu}_t$. As shown in Table \ref{tab:ablation}, the deterministic perturbation largely underperforms the full model. In terms of the objective function, we observe that removing $\mathcal{L}_{MLE}(\theta)$ results in larger performance drops, suggesting that using both augmented and original instance as a single batch is crucial for performance improvement. In addition, the experiment without $D_{KL}$ shows the importance of imposing a constraint on the distribution of perturbation with the KL-term.

% In terms of the objective function, we observe that removing $\:\mathcal{L}_{MLE}(\theta)$ from objectives drops the performance a lot, suggesting that using both augmented and original instance as single batch is important to improve the performance with our data augmentation method. In addition, ablations without $D_{KL}$ shows the importance of the KL divergence to impose a constraint on the distribution of perturbation.

\begin{figure*}[ht!]
    \centering
    \includegraphics[width=1\linewidth]{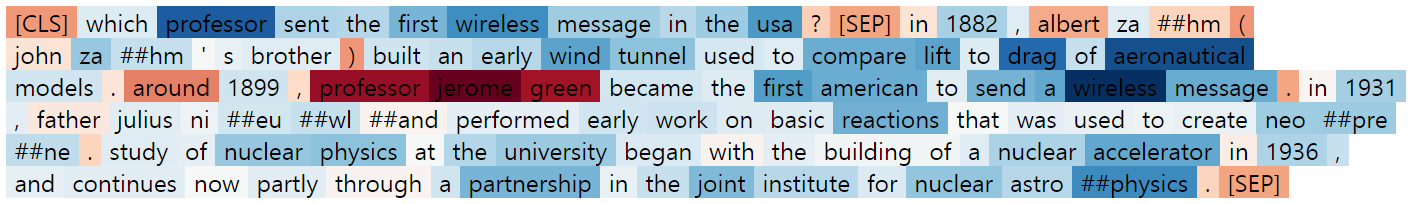}
    \vspace{-0.2in}
    \caption{\small\textbf{Visualization of the Perturbation.} Dark red color indicates the perturbation is near to one, i.e. the corresponding word is rarely perturbed. In contrast, dark blue color indicates the word is relatively more perturbed than others.}
    \label{fig:visualize}
    \vspace{-0.2in}
\end{figure*}

\begin{figure}[ht!]
    \centering
    \includegraphics[width=0.9\linewidth]{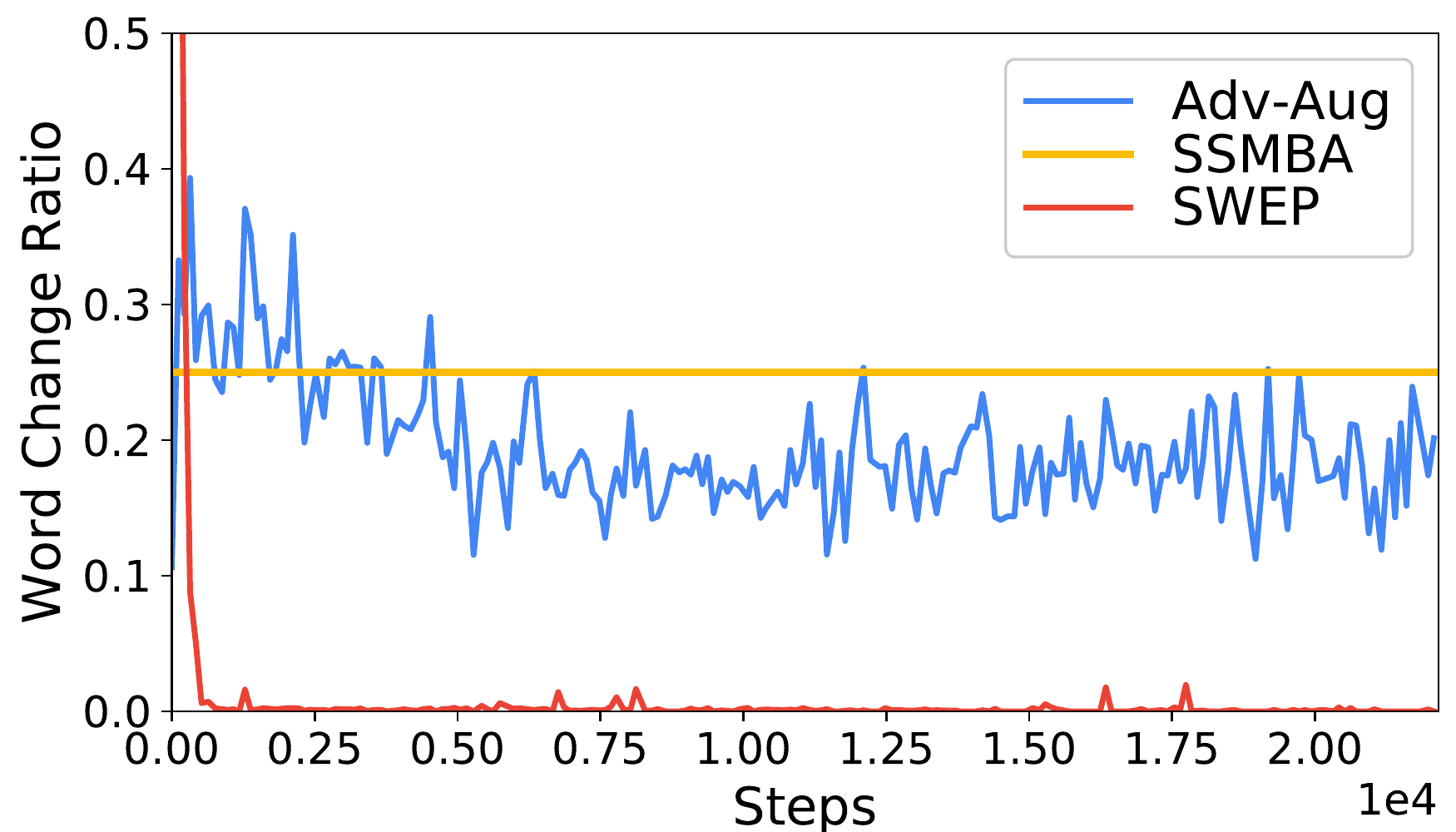}
    \vspace{-0.15in}
    \caption{\small \textbf{Quantitative Analysis.} Plot the extent to how many words changed by perturbation during training.}
    \label{fig:perturb}
    \vspace{-0.2in}
\end{figure}

\subsection{Quantitative Analysis}
We quantitatively analyze the intensity of perturbations given to the input during the training.
To quantitatively measure the semantic drift, we measure the extent to how many words are replaced with another word during training for each data augmentation method and plot it in Fig.~\ref{fig:perturb}. Unlike SSMBA, which replaces the predefined percentage of words with others, the adversarial augmentation (Adv-Aug) or SWEP perturbs the word embeddings in the latent space. We project the perturbed embedding back to the input space to count how many words are changed. Specifically, each word $\rvw_t \in \mathbb{R}^{|\mathcal{V}|}$ is represented as the one-hot vector and mapped to word vector as $\rve_t = W_e \rvw_t$, where $\mathcal{V}$ denotes the vocabulary for training data and $W_e \in \mathbb{R}^{d \times |\mathcal{V}|}$ is the word embedding matrix. Then, the perturbed word embedding $\Tilde{\rve}_t$ is projected back to one-hot vector $\Tilde{\rvw}_t$ as follows:
\begin{equation}
    \begin{gathered}
        (v_1, \ldots, v_d)^\top = W_e^\top \Tilde{\rve}_t \\
        j = \argmax_i\{v_1, \ldots, v_i, \ldots, v_d \} \\
        \Tilde{\rvw}_t = \text{one-hot}(j, |\mathcal{V}|)
    \end{gathered}
\end{equation}
where one-hot($j, |\mathcal{V}|$) makes a one hot vector of which $j$-th component is one with the length $|\mathcal{V}|$. 

In Fig.~\ref{fig:perturb}, we plot the ratio of how many words are replaced with others in raw data before and after each perturbation for each batch as training goes on. In Fig.~\ref{fig:concept}, for example, SSMBA changes about 11 raw words while SWEP does not change any words. We observe that around 20\% of perturbed words are not projected back to each original word if we apply the adversarial augmentation. Also, we see that the adversarial augmentation largely changes the semantics of the words although the perturbation at the final layer is within the epsilon neighborhood of its latent embedding. In contrast, the perturbation by SWEP rarely changes the original words except in the very early stage of training. This observation implies that SWEP learns the range of perturbation that preserves the semantics of the original input, which is important when augmenting data for QA tasks and verifies our concept described in Fig.~\ref{fig:concept}. 

\subsection{Qualitative Analysis}
In Fig.~\ref{fig:visualize}, we visualize the value of the $l_2$ distance between the original word and one with the perturbation after the training. We observe that the perturbation function $q_\phi$ learns to generate adaptive perturbations for each word 
% without a drastic change of semantics of the original input 
(i.e. the lowest intensity of perturbation on answer-like words ``professor jerome green"). However, it is still unknown why the intensity of certain word is higher than the others and how much difference affects the dynamics of training. We have included more observation such as embedding space visualization in Figure~\ref{fig:embedding}. 
% $\frac{1}{d} \sum_{i=1}^d \mu_{t,i}$ for each $t$-th word after the training.

%% file: 5conclusion.tex
\section{Conclusion}
We proposed a simple yet effective data augmentation method based on a stochastic word embedding perturbation for out-of-distribution QA tasks. Specifically, our stochastic noise generator learns to generate the adaptive noise depending on the contextualized embedding of each word. It maximizes the likelihood of input with perturbation, such that it learns to modulate the intensity of perturbation for each word embedding without changing the semantic of the given question and paragraph. We augmented the training data with the perturbed samples using our method, and trained the model with only a single source dataset and evaluate it on datasets from five different domains as well as the in-domain dataset. Based on the experimental results, we verified that our method improves both the performance of in-domain generalization and robustness to distributional shifts, outperforming the baseline data augmentation methods. Further quantitative and qualitative analysis suggest that our method learns to generate adaptive perturbation without a semantic drift.

%% file: 6broader-impact.tex
\section*{Broader Impact}
Our data augmentation method SWEP efficiently improves the robustness of the QA model to unseen out-of-domain data with a few additional computational cost. This robustness is crucial to the success of the real-world QA models, since they frequently encounter questions for unseen domains, from the end-users. While previous works such as  \cite{robustqa} require a set of several heterogeneous datasets to learn domain-invariant representations, such is not a sample-efficient method, while our method is simple yet effective and can improve the robustness of the QA model only when trained on a single source dataset.

\section*{Acknowledgement}
This work was supported by 
Institute of Information \& communications Technology Planning \& Evaluation (IITP) grant funded by the Korea government (MSIT) (No.2019-0-00075, Artificial Intelligence Graduate School Program(KAIST)), 
Samsung Electronics Co., Ltd, 
42Maru,
and the Engineering Research Center Program through the National Research Foundation of Korea (NRF) funded by the Korean Government MSIT (NRF-2018R1A5A1059921).

%% file: 7appendix.tex
\appendix
\section{Experimental Setup}
\subsection{Dataset Statistics}
Table~\ref{datastats} describes detailed dataset statistics.

\subsection{Baselines}
\begin{enumerate}
    \item \textbf{Word-Dropout} We set the same dropout probability as 0.1, which is the same dropout probability of the backbone networks --- BERT, ELECTRA, and T5 model.
    
    \item \textbf{Adv-Aug} We follow the adversarial perturbation from \cite{AdvAug}. We set the number of iteration for perturbation as 5, which is much fewer steps than the original paper due to the computational cost.
    
    \item \textbf{SSMBA} We use the official code of the original paper\footnote{\url{https://github.com/nng555/ssmba}} to augment the training data from SQuAD. We set the probability of masking 0.25 and sample 8 different examples for each training data instance. In total, we synthesize 426,266 additional training instances.

    \item \textbf{Prior-Aug} We set the $\alpha$ as 0.1 which is the dropout probability of the backbone networks.
    
\end{enumerate}

\subsection{Data Augmentation with QG}
Following the experimental setup from \citet{hcvae}, we split the original SQuAD validation dataset by half into new validation and test set.  We download the synthetic QA pairs generated by their generative model Info-HCVAE from the github\footnote{\url{https://github.com/seanie12/Info-HCVAE}} and augment SQuAD training data with them.  They leverage the generative model to sample QA pairs from unlabeled paragraph of HarvestingQA dataset\footnote{\url{https://github.com/xinyadu/HarvestingQA}} \cite{harvesting-qg}, varying the different portion of unlabeled paragraph (denoted as H$\times5\%$-H$\times 50\%$).  We first finetune BERT-base QA model with the synthetic QA pairs generated for 2 epochs and further train it with the original SQuAD training data for another 2 epochs. We use AdamW optimizer \cite{adamw} and set learning rate $2\cdot10^{-5}$ and $3\cdot10^{-5}$ for pretraining and finetuning, respectively with batch size 32. We choose the best checkpoint based on the F1 score from the new validation dataset and evaluate F1 and Exact Match (EM) score on the new test dataset.
\label{appendix-qg}

\begin{table}[t]
	\centering
	\resizebox{0.48\textwidth}{!}{
	\begin{tabular}{llll}
		\midrule[0.8pt]
		\textbf{Datasets} & \textbf{Train (\#)} & \textbf{Valid (\#)} & \textbf{Test (\#)}  \\
		\midrule[0.8pt]
		{SQuAD} & {86,588} & {10,507} & {-}  \\
		\midrule[0.8pt]
		{BioASQ} & {-} &  {-} & {1,504} \\
		{New Wikipedia} & {-} & {-} &{7,938} \\
		{New York Times} & {-} & {-} &{10,065} \\
		{Reddit} & {-} & {-} & {9,803} \\
		{Amazon} & {-} & {-} & {9,885} \\
		\midrule[0.8pt]
		{HarvestQA} & {1,259,691} & {-} & {-}  \\
		\midrule[0.8pt]
	\end{tabular}
	}
	\caption{The statistics and the data source of SQuAD, BioASQ, new Wikipedia, New York Times, Reddit, Amazon, and Harvesting QA.}
	\label{datastats}
\end{table}

\subsection{Computational Cost}
\paragraph{The number of parameters}
Our SWEP model requires few additional learnable parameters relative to the size of the language model. Specifically, our model costs only $3d^2 + 3d$ number of additional parameters, which is less than 2M in the case of BERT-base model where $d=768$. Compared to 110M parameters of BERT-base model, our model does not increase the number of parameters a lot.

\paragraph{Computing infrastructure and Runtime}
In the case of the BERT-base model, the fine-tuning with SWEP costs less than 4 GPU hours with a single Titan XP GPU.

\section{Algorithm}
We describe the whole training procedure described in the section 3.3 as follows:
\renewcommand{\algorithmicrequire}{\textbf{Input:}}
\renewcommand{\algorithmicensure}{\textbf{Output:}}
\begin{algorithm}[h]
    \caption{SWEP}
\begin{algorithmic}[1]
    \STATE \textbf{Input:} \\ Pre-trained Language Model $\theta$ \\ Dataset $\mathcal{D} = \{(\rvx^{(1)}, \rvy^{(1)}), ... , (\rvx^{(N)}, \rvy^{(N)})\}$
    \WHILE {\textit{training}}
    \FOR {$(\rvx^{(i)}, \rvy^{(i)})$ in $\mathcal{D}$}
    \STATE Forward data without perturbation to compute $\log p_\theta (\rvy^{(i)}|\rvx^{(i)})$
    \STATE Sample $\rvz \sim q_\phi(\rvz|\rvx^{(i)})$
    \STATE Forward data with perturbation and compute $\mathcal{L}_{noise}(\phi, \theta)$ 
    \STATE Update $\theta$, $\phi$ with $\mathcal{L}(\phi, \theta)$
    \ENDFOR 
    \ENDWHILE
\end{algorithmic}
\label{algo}
\end{algorithm}

\begin{table*}[ht]
	\centering
	\resizebox{0.8\textwidth}{!}{
	\begin{tabular}{lrrrr}
		\toprule
		\textbf{Word Frequency Rank} & [1, 100] & (100, 500] & (500, 5K] & (5K, 10K]  \\
		\midrule[0.8pt]
		\midrule[0.8pt]
		{$k$-NN $l_2$-dist. ($k=5$)} & {0.6618} & {0.7893} & {0.8474} & {0.8973}  \\
		\midrule[0.8pt]
		{$l_2$-dist. between before / after perturb.} & {0.2386} &  {0.2989} & {0.3542} & {0.4099} \\
		{Mean $\mu$} & {1.2153} &  {1.2402} & {1.2495} & {1.2543} \\
		\bottomrule
	\end{tabular}
	}
	\caption{The $l_2$ distance of $k$-NN nearest neighbor, $l_2$ distance between embeddings before and after perturbation, and the average $\mu$ value  of the word embedding from BERT, segmented by the word frequency rank (Lower rank indicates high-frequency word).}
	\label{tab:analysis}
\end{table*}

\section{Further Analysis}
Motivated by observations from~\citep{SentEmb}, we further analyze the adaptive perturbation for each word. \citet{SentEmb} observe that low-frequency words disperse sparsely while high-frequency words concentrate densely on the word embedding space of BERT. Following the setting of~\citep{SentEmb}, we first measure the $l_2$ distance between $k$-nearest neighbors of each word embedding. Specifically, we rank each word (wordpiece tokens) by frequency counted based on the SQuAD train set and sample 100 examples from the SQuAD train set for analysis. In Table~\ref{tab:analysis}, we also observe that low-frequency words have more distance to their neighbor than high-frequency words. 
Then, we measure the average $l_2$ distance of word embedding before and after perturbation and the average perturbation size for each word as $\frac{1}{d} \sum_{i=1}^d \mu_{t,i}$ after the training. We observe that low-frequency words tend to be perturbed more than high-frequency words. This observation suggests that the noise generator can recognize acceptable extents to perturb words depend on the word embedding distribution then tends to generate more perturbation on sparsely dispersed low-frequency words and less perturbation on densely concentrated high-frequency words. Note that we use beta annealing to magnify the difference for analysis so that the $\beta$ becomes zero in the second epoch.

\section{Embedding Space Visualization}
\begin{figure*}[h]
    \centering
    \includegraphics[width=0.9\linewidth]{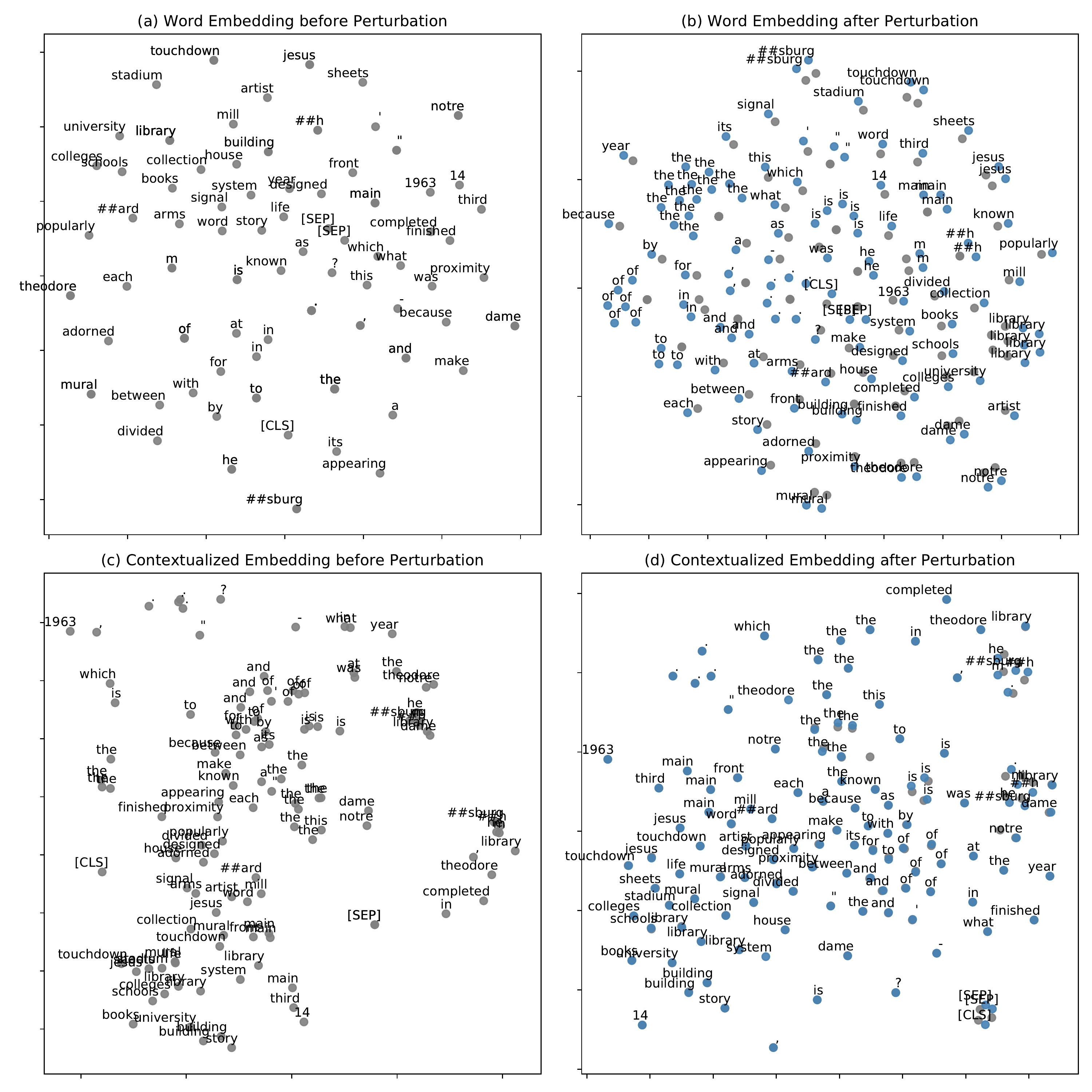}
    \caption{\textbf{Visualization.} Overview of how the input is perturbed with SWEP. Contextualized embedding indicates the hidden states from the last layer of transformers. Blue points indicate embeddings after perturbation.}
    \label{fig:embedding}
\end{figure*}

In Figure~\ref{fig:embedding}, we visualize the embedding space using t-SNE~\citep{tsne} for both word embedding ((a), (b)) and contextualized embedding ((c), (d)) before and after perturbation from ELECTRA-small model. We sample the example from the SQuAD training set, which is the same example as Figure 1 in the main paper. SWEP encodes each input tokens $x_t$ to hidden representation $\rvh_t$ with transformers and outputs a desirable noise for each word embedding. The noise $\rvz_t$ is multiplied with the word embedding $\rve_t$ of each token $x_t$. We observe that the perturbed word embedding is mapped to a different space against original word embedding, however, the contextualized embedding is not much changed by the perturbation. Note that absolute positions are different in each plot because of the randomness inherent in the t-SNE algorithms.

%% file: main.bbl
\begin{thebibliography}{33}
\expandafter\ifx\csname natexlab\endcsname\relax\def\natexlab#1{#1}\fi

\bibitem[{Balaji et~al.(2018)Balaji, Sankaranarayanan, and
  Chellappa}]{meta-reg}
Yogesh Balaji, Swami Sankaranarayanan, and Rama Chellappa. 2018.
\newblock Metareg: Towards domain generalization using meta-regularization.
\newblock \emph{Advances in Neural Information Processing Systems},
  31:998--1008.

\bibitem[{Clark et~al.(2020)Clark, Luong, Le, and Manning}]{Electra}
Kevin Clark, Minh{-}Thang Luong, Quoc~V. Le, and Christopher~D. Manning. 2020.
\newblock {ELECTRA:} pre-training text encoders as discriminators rather than
  generators.
\newblock In \emph{8th International Conference on Learning Representations,
  {ICLR} 2020, Addis Ababa, Ethiopia, April 26-30, 2020}.

\bibitem[{Devlin et~al.(2019)Devlin, Chang, Lee, and Toutanova}]{BERT}
Jacob Devlin, Ming{-}Wei Chang, Kenton Lee, and Kristina Toutanova. 2019.
\newblock {BERT:} pre-training of deep bidirectional transformers for language
  understanding.
\newblock In \emph{Proceedings of the 2019 Conference of the North American
  Chapter of the Association for Computational Linguistics: Human Language
  Technologies, {NAACL-HLT} 2019, Minneapolis, MN, USA, June 2-7, 2019, Volume
  1 (Long and Short Papers)}, pages 4171--4186.

\bibitem[{Du and Cardie(2018)}]{harvesting-qg}
Xinya Du and Claire Cardie. 2018.
\newblock Harvesting paragraph-level question-answer pairs from wikipedia.
\newblock In \emph{Proceedings of the 56th Annual Meeting of the Association
  for Computational Linguistics (Volume 1: Long Papers)}, pages 1907--1917.

\bibitem[{Fisch et~al.(2019)Fisch, Talmor, Jia, Seo, Choi, and Chen}]{mrqa}
Adam Fisch, Alon Talmor, Robin Jia, Minjoon Seo, Eunsol Choi, and Danqi Chen.
  2019.
\newblock {MRQA} 2019 shared task: Evaluating generalization in reading
  comprehension.
\newblock In \emph{Proceedings of the 2nd Workshop on Machine Reading for
  Question Answering, MRQA@EMNLP 2019, Hong Kong, China, November 4, 2019},
  pages 1--13.

\bibitem[{Jia and Liang(2017)}]{adversarial-squad}
Robin Jia and Percy Liang. 2017.
\newblock Adversarial examples for evaluating reading comprehension systems.
\newblock In \emph{Proceedings of the 2017 Conference on Empirical Methods in
  Natural Language Processing, {EMNLP} 2017, Copenhagen, Denmark, September
  9-11, 2017}, pages 2021--2031.

\bibitem[{Kingma and Welling(2014)}]{vae}
Diederik~P. Kingma and Max Welling. 2014.
\newblock Auto-encoding variational bayes.
\newblock In \emph{International Conference on Learning Representations, {ICLR}
  2014,}.

\bibitem[{Krizhevsky et~al.(2012)Krizhevsky, Sutskever, and Hinton}]{ImageNet}
Alex Krizhevsky, Ilya Sutskever, and Geoffrey~E. Hinton. 2012.
\newblock Imagenet classification with deep convolutional neural networks.
\newblock In \emph{Advances in Neural Information Processing Systems 25: 26th
  Annual Conference on Neural Information Processing Systems 2012. Proceedings
  of a meeting held December 3-6, 2012, Lake Tahoe, Nevada, United States}.

\bibitem[{Lee et~al.(2020)Lee, Lee, Jeong, Kim, and Hwang}]{hcvae}
Dong~Bok Lee, Seanie Lee, Woo~Tae Jeong, Donghwan Kim, and Sung~Ju Hwang. 2020.
\newblock Generating diverse and consistent {QA} pairs from contexts with
  information-maximizing hierarchical conditional vaes.
\newblock In \emph{Proceedings of the 58th Annual Meeting of the Association
  for Computational Linguistics, {ACL} 2020, Online, July 5-10, 2020}, pages
  208--224.

\bibitem[{Lee et~al.(2019)Lee, Kim, and Park}]{robustqa}
Seanie Lee, Donggyu Kim, and Jangwon Park. 2019.
\newblock Domain-agnostic question-answering with adversarial training.
\newblock In \emph{Proceedings of the 2nd Workshop on Machine Reading for
  Question Answering, MRQA@EMNLP 2019, Hong Kong, China, November 4, 2019},
  pages 196--202.

\bibitem[{Li et~al.(2020)Li, Zhou, He, Wang, Yang, and Li}]{SentEmb}
Bohan Li, Hao Zhou, Junxian He, Mingxuan Wang, Yiming Yang, and Lei Li. 2020.
\newblock On the sentence embeddings from pre-trained language models.
\newblock In \emph{Proceedings of the 2020 Conference on Empirical Methods in
  Natural Language Processing, {EMNLP} 2020, Online, November 16-20, 2020},
  pages 9119--9130.

\bibitem[{Li et~al.(2018)Li, Yang, Song, and Hospedales}]{l2g}
Da~Li, Yongxin Yang, Yi-Zhe Song, and Timothy Hospedales. 2018.
\newblock Learning to generalize: Meta-learning for domain generalization.
\newblock In \emph{Proceedings of the AAAI Conference on Artificial
  Intelligence}.

\bibitem[{Loshchilov and Hutter(2019)}]{adamw}
Ilya Loshchilov and Frank Hutter. 2019.
\newblock Decoupled weight decay regularization.
\newblock In \emph{International Conference on Learning Representations}.

\bibitem[{Maaten and Hinton(2008)}]{tsne}
L.~V.~D. Maaten and Geoffrey~E. Hinton. 2008.
\newblock Visualizing data using t-sne.
\newblock \emph{Journal of Machine Learning Research}, 9:2579--2605.

\bibitem[{Miller et~al.(2020)Miller, Krauth, Recht, and
  Schmidt}]{natural-shift}
John Miller, Karl Krauth, Benjamin Recht, and Ludwig Schmidt. 2020.
\newblock The effect of natural distribution shift on question answering
  models.
\newblock In \emph{International Conference on Machine Learning}, pages
  6905--6916. PMLR.

\bibitem[{Ng et~al.(2020)Ng, Cho, and Ghassemi}]{SSMBA}
Nathan Ng, Kyunghyun Cho, and Marzyeh Ghassemi. 2020.
\newblock Ssmba: Self-supervised manifold based data augmentation for improving
  out-of-domain robustness.
\newblock In \emph{Proceedings of the 2020 Conference on Empirical Methods in
  Natural Language Processing (EMNLP)}, pages 1268--1283.

\bibitem[{Pham et~al.(2021)Pham, Wang, Yang, and Neubig}]{meta-backtranslation}
Hieu Pham, Xinyi Wang, Yiming Yang, and Graham Neubig. 2021.
\newblock Meta back-translation.
\newblock In \emph{International Conference on Learning Representations}.

\bibitem[{Raffel et~al.(2020)Raffel, Shazeer, Roberts, Lee, Narang, Matena,
  Zhou, Li, and Liu}]{T5}
Colin Raffel, Noam Shazeer, Adam Roberts, Katherine Lee, Sharan Narang, Michael
  Matena, Yanqi Zhou, Wei Li, and Peter~J. Liu. 2020.
\newblock Exploring the limits of transfer learning with a unified text-to-text
  transformer.
\newblock \emph{Journal of Machine Learning Research}, 21(140):1--67.

\bibitem[{Rajpurkar et~al.(2016)Rajpurkar, Zhang, Lopyrev, and Liang}]{SQuAD}
Pranav Rajpurkar, Jian Zhang, Konstantin Lopyrev, and Percy Liang. 2016.
\newblock Squad: 100, 000+ questions for machine comprehension of text.
\newblock In \emph{Proceedings of the 2016 Conference on Empirical Methods in
  Natural Language Processing, {EMNLP} 2016, Austin, Texas, USA, November 1-4,
  2016}, pages 2383--2392.

\bibitem[{Sennrich et~al.(2016)Sennrich, Haddow, and Birch}]{word-dropout}
Rico Sennrich, Barry Haddow, and Alexandra Birch. 2016.
\newblock Edinburgh neural machine translation systems for wmt 16.
\newblock In \emph{Proceedings of the First Conference on Machine Translation:
  Volume 2, Shared Task Papers}.

\bibitem[{Seo et~al.(2017)Seo, Kembhavi, Farhadi, and Hajishirzi}]{bidaf}
Min~Joon Seo, Aniruddha Kembhavi, Ali Farhadi, and Hannaneh Hajishirzi. 2017.
\newblock Bidirectional attention flow for machine comprehension.
\newblock In \emph{International Conference on Learning Representations, {ICLR}
  2017,}.

\bibitem[{Shazeer and Stern(2018)}]{adafactor}
Noam Shazeer and Mitchell Stern. 2018.
\newblock Adafactor: Adaptive learning rates with sublinear memory cost.
\newblock In \emph{International Conference on Machine Learning}.

\bibitem[{Srivastava et~al.(2014)Srivastava, Hinton, Krizhevsky, Sutskever, and
  Salakhutdinov}]{dropout}
Nitish Srivastava, Geoffrey Hinton, Alex Krizhevsky, Ilya Sutskever, and Ruslan
  Salakhutdinov. 2014.
\newblock Dropout: a simple way to prevent neural networks from overfitting.
\newblock \emph{The journal of machine learning research}.

\bibitem[{Tsatsaronis et~al.(2012)Tsatsaronis, Schroeder, Paliouras,
  Almirantis, Androutsopoulos, Gaussier, Gallinari, Arti{\`{e}}res, Alvers,
  Zschunke, and Ngomo}]{bioasq}
George Tsatsaronis, Michael Schroeder, Georgios Paliouras, Yannis Almirantis,
  Ion Androutsopoulos, {\'{E}}ric Gaussier, Patrick Gallinari, Thierry
  Arti{\`{e}}res, Michael~R. Alvers, Matthias Zschunke, and
  Axel{-}Cyrille~Ngonga Ngomo. 2012.
\newblock Bioasq: {A} challenge on large-scale biomedical semantic indexing and
  question answering.
\newblock In \emph{Information Retrieval and Knowledge Discovery in Biomedical
  Text, Papers from the 2012 {AAAI} Fall Symposium}.

\bibitem[{Tseng et~al.(2020)Tseng, Lee, Huang, and Yang}]{cross-domain}
Hung-Yu Tseng, Hsin-Ying Lee, Jia-Bin Huang, and Ming-Hsuan Yang. 2020.
\newblock Cross-domain few-shot classification via learned feature-wise
  transformation.
\newblock In \emph{International Conference on Learning Representations}.

\bibitem[{Vaswani et~al.(2017)Vaswani, Shazeer, Parmar, Uszkoreit, Jones,
  Gomez, Kaiser, and Polosukhin}]{transformer}
Ashish Vaswani, Noam Shazeer, Niki Parmar, Jakob Uszkoreit, Llion Jones,
  Aidan~N. Gomez, Lukasz Kaiser, and Illia Polosukhin. 2017.
\newblock Attention is all you need.
\newblock In \emph{Advances in Neural Information Processing Systems 30: Annual
  Conference on Neural Information Processing Systems 2017, 4-9 December 2017,
  Long Beach, CA, {USA}}, pages 5998--6008.

\bibitem[{Verma et~al.(2019{\natexlab{a}})Verma, Lamb, Beckham, Najafi,
  Mitliagkas, Lopez{-}Paz, and Bengio}]{mixup}
Vikas Verma, Alex Lamb, Christopher Beckham, Amir Najafi, Ioannis Mitliagkas,
  David Lopez{-}Paz, and Yoshua Bengio. 2019{\natexlab{a}}.
\newblock Manifold mixup: Better representations by interpolating hidden
  states.
\newblock In \emph{Proceedings of the 36th International Conference on Machine
  Learning, {ICML} 2019, 9-15 June 2019, Long Beach, California, {USA}}, pages
  6438--6447.

\bibitem[{Verma et~al.(2019{\natexlab{b}})Verma, Lamb, Beckham, Najafi,
  Mitliagkas, Lopez-Paz, and Bengio}]{manifold-mixup}
Vikas Verma, Alex Lamb, Christopher Beckham, Amir Najafi, Ioannis Mitliagkas,
  David Lopez-Paz, and Yoshua Bengio. 2019{\natexlab{b}}.
\newblock Manifold mixup: Better representations by interpolating hidden
  states.
\newblock In \emph{International Conference on Machine Learning}. PMLR.

\bibitem[{Volpi et~al.(2018)Volpi, Namkoong, Sener, Duchi, Murino, and
  Savarese}]{AdvAug}
Riccardo Volpi, Hongseok Namkoong, Ozan Sener, John~C. Duchi, Vittorio Murino,
  and Silvio Savarese. 2018.
\newblock Generalizing to unseen domains via adversarial data augmentation.
\newblock In \emph{Advances in Neural Information Processing Systems 31: Annual
  Conference on Neural Information Processing Systems 2018, NeurIPS 2018,
  December 3-8, 2018, Montr{\'{e}}al, Canada}, pages 5339--5349.

\bibitem[{Wei and Zou(2019)}]{EDA}
Jason~W. Wei and Kai Zou. 2019.
\newblock {EDA:} easy data augmentation techniques for boosting performance on
  text classification tasks.
\newblock In \emph{Proceedings of the 2019 Conference on Empirical Methods in
  Natural Language Processing and the 9th International Joint Conference on
  Natural Language Processing, {EMNLP-IJCNLP} 2019, Hong Kong, China, November
  3-7, 2019}, pages 6381--6387.

\bibitem[{Xie et~al.(2020)Xie, Dai, Hovy, Luong, and Le}]{UDA}
Qizhe Xie, Zihang Dai, Eduard~H. Hovy, Thang Luong, and Quoc Le. 2020.
\newblock Unsupervised data augmentation for consistency training.
\newblock In \emph{Advances in Neural Information Processing Systems 33: Annual
  Conference on Neural Information Processing Systems 2020, NeurIPS 2020,
  December 6-12, 2020, virtual}.

\bibitem[{Yun et~al.(2019)Yun, Han, Oh, Chun, Choe, and Yoo}]{cutmix}
Sangdoo Yun, Dongyoon Han, Seong~Joon Oh, Sanghyuk Chun, Junsuk Choe, and
  Youngjoon Yoo. 2019.
\newblock Cutmix: Regularization strategy to train strong classifiers with
  localizable features.
\newblock In \emph{Proceedings of the IEEE International Conference on Computer
  Vision}.

\bibitem[{Zhang and Bansal(2019)}]{semantic-drift}
Shiyue Zhang and Mohit Bansal. 2019.
\newblock Addressing semantic drift in question generation for semi-supervised
  question answering.
\newblock In \emph{Proceedings of the 2019 Conference on Empirical Methods in
  Natural Language Processing and the 9th International Joint Conference on
  Natural Language Processing (EMNLP-IJCNLP)}, pages 2495--2509.

\end{thebibliography}
